\documentclass[conference]{IEEEtran}
\usepackage{cite}
\usepackage{amsmath,amssymb,amsfonts}
\usepackage{algorithmic}
\usepackage{graphicx}
\usepackage{tabularx}
\usepackage{booktabs}
\usepackage{url}
\usepackage{multirow}
\usepackage{textcomp}
\usepackage{xcolor}
\def\BibTeX{{\rm B\kern-.05em{\sc i\kern-.025em b}\kern-.08em
    T\kern-.1667em\lower.7ex\hbox{E}\kern-.125emX}}

\usepackage{amsthm}

\theoremstyle{definition}
\newtheorem{example}{Example}

\begin{document}

\title{From LLM-Generated Specifications to Learned Quadruped Locomotion}

\author{
\IEEEauthorblockN{
Merve Atasever\IEEEauthorrefmark{1},
Keyan Azbijari\IEEEauthorrefmark{1},
Cagan Bakirci\IEEEauthorrefmark{1},
Alfredo Reina Corona\IEEEauthorrefmark{1}, \\
Tolga Izdas\IEEEauthorrefmark{1},
Richard Yang\IEEEauthorrefmark{2},
Erdem Biyik\IEEEauthorrefmark{1},
Jyotirmoy V. Deshmukh\IEEEauthorrefmark{1}
}

\IEEEauthorblockA{\IEEEauthorrefmark{1}\textit{Department of Computer Science} \\
\textit{University of Southern California} \\
Los Angeles, CA, USA \\
\{atasever, azbijari, cbakirci, reinacor, izdas, biyik, jdeshmuk\}@usc.edu
}

\IEEEauthorblockA{\IEEEauthorrefmark{2}\textit{University of Florida} \\
Gainesville, FL, USA \\ 
yangrichard@ufl.edu
}
}

\maketitle

\begin{abstract}
Quadruped robot locomotion policies are often trained using reinforcement learning, which
in turn relies heavily on hand-crafted reward functions. Designing reward functions requires substantial manual engineering, and it is often unclear which local rewards will induce the desired global behavior. Shaped rewards from formal specifications in languages like Signal Temporal Logic (STL) can make rewards more interpretable, but writing STL specifications itself still requires domain expertise. We study whether large language models (LLMs) can fill this gap by generating Parametric Signal Temporal Logic (PSTL) specifications that are subsequently used for policy learning. Given a natural language locomotion objective and a constrained specification grammar, GPT-5.5 and Qwen 3.6 independently propose STL templates for command tracking, safety, and gait structure. We instantiate the parameters of the generated PSTL templates using expert trajectories and retain only specifications that are consistent with demonstrated expert behavior. The resulting specifications are then transformed into smooth, finite-history reward functions and used to train a quadruped locomotion policy with Proximal Policy Optimization (PPO) in MuJoCo XLA (MJX). We evaluate both \emph{gait-aware} and \emph{gait-agnostic} settings. The former specifies walking-trot, trot, and bound regimes, while the latter allows contact patterns to emerge from the task objective. We compare against hand-engineered rewards, Text2Reward-style LLM-generated reward code, and an expert-switching oracle. Gait-aware Qwen 3.6 specifications achieved 100\% survival and command success across all tested speeds (0.3--2.1 m/s) and matched the target gait at high speeds, whereas Text2Reward achieved 0\% for both metrics at $\geq 1.9$ m/s. Videos: \url{https://stl-locomotion.github.io/}
\end{abstract}


\section{Introduction}

Deep RL has enabled quadruped robots to learn agile locomotion from simulations and expert demonstrations, but performance still depends strongly on reward engineering \cite{dayan2002reward,tan2018sim,learningquadsci,eschmann2021reward,icarte2022reward,caluwaerts2023barkour,zakka2025mujocoplayground}. A typical locomotion reward is a weighted sum of instantaneous tracking, posture, smoothness, contact, and energy terms. Although such rewards can produce performant policies, selecting various reward components and coefficients requires substantial domain knowledge and empirical tuning, with design choices often justified through ablations rather than first principles \cite{hare2019dealing,eschmann2021reward,kim2025learning}. Moreover, the resulting numerical, local rewards do not explicitly describe the global temporal behavior it is intended to induce. This is especially problematic for multi-gait locomotion: walking, trotting, and bounding differ not only in speed, but also in contact timing and support patterns. 
Alternatives such as inverse RL and preference-based RL reduce reliance on manually specified rewards by inferring objectives from demonstrations or human feedback, but introduce an additional reward-learning or optimization stage \cite{fu2017learning,inverserl,arora2021survey,rlhf,youm2023imitating,li2023fastmimic}. Our approach also uses expert trajectories, but only to calibrate and validate an explicit set of symbolic, global specifications whose structure is proposed by an LLM.

Another approach is to describe desired behavior explicitly using formal logic and convert specification-satisfaction into a learning signal. Prior work on logic-guided RL has leveraged Linear Temporal Logic (LTL) and Signal Temporal Logic (STL) for reward shaping and policy learning, while also exploring the inference of temporal-logic specifications from demonstrations, particularly in navigation and manipulation tasks. \cite{hasanbeig2020deep,li2017reinforcement,liao2020survey,stlrl2,stlrl,stlrl3}. Formal specifications have similarly been incorporated into planning and control for legged robots, including locomotion over cluttered terrain, stability constraints, and STL-informed control for bipedal walking and push recovery \cite{feng20133d,bipedal,audren2016stability,gu2024walking,gu2025robust}. For quadrupeds, however, comparable continuous temporal-logic formulations remain limited. Existing logic-driven gait-learning approaches largely employ discrete logical rules or reward machines defined over foot-contact propositions and operating primarily on contact-level logic and step-wise signals \cite{gaitstrategiesnature,defazio2024learning}.

Signal Temporal Logic or STL \cite{maler2004monitoring} is well suited to this problem because it expresses bounded time-horizon properties directly over real-valued trajectories and associates each specification with a quantitative robustness value measuring its margin of satisfaction or violation \cite{donze2010robust,fainekos2009robustness}. These quantitative semantics makes it possible to use STL not merely as a {\em post hoc} verifier, but as a dense reward signal for RL \cite{li2017reinforcement,balakrishnan2019structured,aksaray2016q,stlrl2}. Unlike conventional rewards, an STL specification can explicitly encode temporal relationships such as how long a condition should persist, how events should be ordered, or how contact patterns should evolve over a finite horizon.

In our previous work, we manually defined parametric STL templates for quadruped safety, command tracking, and gait behavior and calibrated their numerical thresholds from expert rollouts \cite{atasever2026learning}. While
this formulation improves interpretability and gait control relative to heuristic
rewards, it leaves the burden of manual specification design unresolved.

Recent LLM-based reward-design methods address a closely related problem by generating executable dense reward functions directly from natural-language task descriptions \cite{text2reward,ma2024eureka}. In parallel, language models have increasingly been used to translate natural language requirements into temporal logic representations \cite{chen2023nl2tl}. We combine these ideas differently. Instead of asking an LLM to produce the numerical reward itself, we ask it to propose a symbolic temporal specification. Expert trajectories are then used to fit the specification parameters and reject candidates that disagree with demonstrated behavior. The retained specifications are subsequently converted into dense learning signals through STL robustness.

We investigate this design using Google's Barkour quadruped. GPT-5.5 and Qwen 3.6 are prompted with the same locomotion objective and STL grammar, but are not provided numerical thresholds \cite{openai2026gpt55,qwen2026qwen36}. Each model proposes Parametric STL (PSTL) specifications. We fit the values of these parameters from expert trajectories and retain only specifications
that are consistent with those trajectories. During policy learning, we compute smooth trailing-window STL robustness online and use it as a PPO reward \cite{ppo, kohl2004policy}. We evaluate the approach in both \emph{gait-aware} (multi-gait) locomotion, where speed regimes activate gait-specific specifications, and \emph{gait-agnostic} locomotion, where no gait is prescribed and the robot is free to discover its own contact pattern. 

\noindent\textbf{Contributions.}
\begin{enumerate}
\item We introduce a pipeline in which LLMs generate the \emph{structure} of interpretable locomotion specifications while expert data grounds their numerical parameters.
\item We propose a simple expert-consistency filter: a generated specification is discarded when its median robustness on expert data is negative, preventing systematically violated LLM suggestions from entering the reward.
\item We evaluate both gait-aware and gait-agnostic formulations to study how explicitly prescribing gait structure affects the learned locomotion behavior.
\item We benchmark against a heuristic reward, Text2Reward, and an expert-switching oracle.
\end{enumerate}

\begin{figure}[h]
\centering
\includegraphics[width=0.95\linewidth]{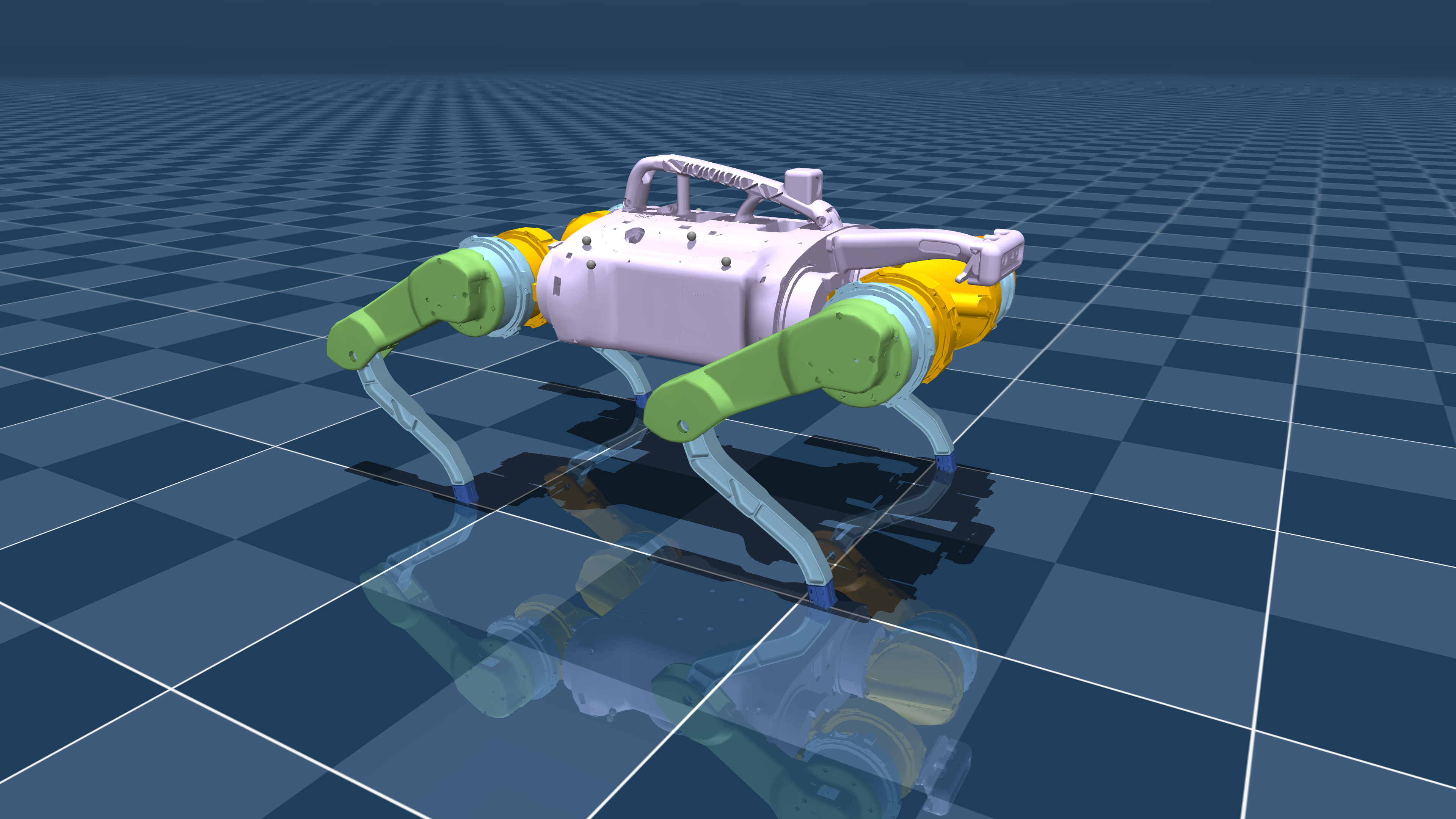}
\caption{Barkour vb robot in MJX.}
\label{fig:barkour}
\end{figure}

\section{Background and Problem Setting}

We consider Barkour vb in MJX \cite{caluwaerts2023barkour,todorov2012mujoco, zakka2025mujocoplayground}. At each control step ($\Delta t=0.02$ s), the policy receives commanded linear and angular velocities $(v_x^{cmd},v_y^{cmd},\omega_z^{cmd})$, proprioceptive observations, the previous action, and a short observation history, and outputs 12 normalized joint-position commands. A PD controller maps these commands to actuator torques. The goal is to learn a policy $\pi_\theta(a_t\mid o_t)$ that tracks commands while remaining stable and energy efficient.

An STL predicate has the form $f(x(t))\ge 0$. Boolean operators combine predicates and temporal operators $\mathbf{G}_{[a,b]}$ and $\mathbf{F}_{[a,b]}$ respectively constrain predicate satisfaction to be globally true over the time interval $[a,b]$ or true at some point during the interval. Quantitative semantics assign a robustness $\rho^\varphi(x,t)\in\mathbb{R}$ to formula $\varphi$: positive values indicate satisfaction, and negative values indicate violation \cite{donze2010robust,fainekos2009robustness}. Parametric STL (PSTL) replaces numerical constants in predicates or temporal intervals by parameters that can be fit from data \cite{asarin2011parametric}.

We evaluate each temporal specification over the trailing window $\mathcal{W}_t=[t-H,t]$, and the resulting specification-level robustness values are subsequently combined using the smooth aggregation described in the next section \cite{gilpin2020smooth}.

\section{LLM-Generated STL Reward Pipeline}

\subsection{Specification Generation}

The LLM receives (i) a natural-language description of Barkour locomotion and (ii) an allowed STL grammar. We explicitly ask for \emph{template structure only}: thresholds and temporal constants remain symbolic so that they can be estimated from expert data rather than invented by the language model. GPT-5.5 and Qwen 3.6 are queried independently with the same prompt.

For the multi-gait setting, the prompt describes three forward-speed regimes: walking-trot ($0\le v_x^{cmd}\le 0.7$ m/s), trot ($0.7< v_x^{cmd} \le 1.7$ m/s), and bound ($v_x^{cmd}>1.7$ m/s). Generated formulas can refer to shared tracking/safety signals and regime-specific contact statistics. Typical template families include examples shown below\footnote{Complete sets of generated specifications and prompts are available on the project website: \url{https://stl-locomotion.github.io/}}. 

Let the trailing evaluation window at time $t$ be defined as
\begin{equation}
\mathcal{W}_t = \{t-H,\ldots,t\},
\end{equation}
where $H$ denotes the robustness horizon. For an STL formula of the form
$\mathbf{G}_{\mathcal{W}}\mu$, where $\mu$ is a predicate, the robustness at
time $t$ is computed as
\begin{equation}
\rho^{\mathbf{G}_{\mathcal{W}}\mu}(t)
=
\min_{k\in\mathcal{W}_t} \rho^{\mu}(k).
\end{equation}
Thus, the robustness is determined by the worst predicate margin observed
within the trailing temporal window.

\begin{example}
Torque limits: For Barkour vb, the
actuators provide a peak output torque of $18$~Nm ($\tau^{\max}$) at each joint,
hence motor torque limits must be bounded by this limit \cite{caluwaerts2023barkour}.
\begin{equation}
\varphi_{\tau}
=
\mathbf{G}_{\mathcal{W}}
\left(
\bigwedge_{j=1}^{12}
|\tau_j| \leq \tau^{\max}
\right).
\end{equation}
The corresponding robustness is
\begin{equation}
\rho_{\tau}(t)
=
\min_{k\in\mathcal{W}_t}
\min_{j\in\{1,\ldots,12\}}
\left(
\tau^{\max} - |\tau_j(k)|
\right).
\end{equation}
\end{example}

\begin{example}
Torso orientation: Roll and pitch are constrained using
\begin{align}
\varphi_{\phi}
&=
\mathbf{G}_{\mathcal{W}}
\left(
|\phi| \leq \phi_{\max}
\right), \\
\varphi_{\theta}
&=
\mathbf{G}_{\mathcal{W}}
\left(
|\theta| \leq \theta_{\max}
\right).
\end{align}
\end{example}

\begin{example}
Center-of-mass (CoM) height:
A minimum center-of-mass height is specified as
\begin{equation}
\varphi_{z}
=
\mathbf{G}_{\mathcal{W}}
\left(
z_{\mathrm{com}} \geq z_{\min}
\right).
\end{equation}
\end{example}

\begin{example}
Support contacts:
To require a minimum number of feet in contact with the ground, we use
\begin{equation}
\varphi_{c}
=
\mathbf{G}_{\mathcal{W}}
\left(
n_c \geq n_{\min}
\right),
\end{equation}
where $n_c$ denotes the number of active foot contacts. 
\end{example}

\begin{example}
Linear velocity tracking:
For each commanded linear velocity component $i\in\{x,y\}$, tracking is
specified as
\begin{equation}
\varphi_{v_i}
=
\mathbf{G}_{\mathcal{W}}
\left(
|v_i - v_i^{\mathrm{cmd}}|
\leq
\epsilon_{v,i}
\right).
\end{equation}
\end{example}

The numerical parameters $\phi_{\max}$, $\theta_{\max}$, $z_{\min}$,
$n_{\min}$, and $\epsilon_{v,i}$
are not generated by the LLM. The LLM determines only the symbolic structure of each STL specification, while the numerical parameters are
subsequently estimated from expert trajectories. The resulting specification-level robustness values are then aggregated to construct the reward used for PPO training.\\

\noindent{\bf Multi-Gait vs Gait-Agnostic Policies}
Using the same gait for low speed and high speed motion may be inefficient or infeasible for quadrupeds \cite{fu2021minimizing, yang2022fast}. To determine transition velocity thresholds between speed regimes, we leverage the Froude number which is used to characterize gait transition speeds of different-sized quadrupeds:
\(
\mathrm{Fr} = \frac{v^2}{g h},
\)
where Fr is the Froude number, $v$ is the linear velocity of the robot's torso, $g$ is the acceleration due to gravity, and $h$ is the maximum leg length \cite{alexander1983dynamic}. 
Building on bio-inspired robotics gait design methodologies and reported gait structures that Barkour vb achieved to succeed, we leverage three speed regimes (modes): walking-trot, trot, and bound, selected based on commanded forward velocity $|v_x^{cmd}(t)|$. Using the Barkour robot's leg length of $0.41~\mathrm{m}$ and the reference Froude numbers of $0.15$ and $0.74$ \cite{humphreys2023bio}, we establish the respective gait transition velocity thresholds at $0.7~\mathrm{m/s}$ and $1.7~\mathrm{m/s}$. Here, diagonal legs move together for mid-speed regime (trot), whereas forelegs make contact approximately in phase, followed by the hind legs to achieve high speeds (bound).

While the multi-gait setting aims to learn a single policy that adapts its gait structure across speed regimes (walking-trot $\rightarrow$ trot $\rightarrow$ bound), the gait-agnostic setting does not enforce any target gait. We therefore use separate prompts for the two settings.

\subsection{Robustness Reward}

For active retained specifications, we aggregate multiple robustness terms using the soft-min
$
\operatorname{smin}^{\mathrm{sp}}_{\beta}(z_1,\dots,z_K)
:= \sum_{k=1}^{K}\pi_k z_k$, where 
$
\pi_k = \frac{e^{-\beta z_k}}{\sum_{j=1}^{K} e^{-\beta z_j}}.
$
For the final reward shaping, each grouped robustness is normalized as $\tanh(\mu)$
to map to a value in $[-1,1]$, preventing large robustness magnitudes from dominating the reward.
The final scalar reward is:
\begin{equation}
r_t
=
w_{\mathrm{safe}}\, \tanh(\mu_{\mathrm{sa}})
+
w_{\mathrm{track}}\, \tanh(\mu_{\mathrm{tr}})
+ 
w_{\mathrm{pattern}}\, \tanh(\mu_{\mathrm{pa}})
\end{equation}

This decomposition makes the generated reward inspectable: during an evaluation we can report both the scalar RL reward and the robustness of each retained formula. 

\subsection{Data Grounding and Expert-Consistency Filtering}

We collect 50 expert trajectories per regime, each containing $500$ time steps, and compute the signals required by every generated template. Numerical parameters for PSTL templates are estimated from expert data statistics using quantiles appropriate to the predicate direction: upper quantiles for error ceilings and lower quantiles for safety floors. 

Moreover, LLMs can propose technically sound but incompatible or overly strict requirements. We take advantage of expert trajectories to filter these specifications. Let
\begin{equation}
R_\varphi =
\left\{
\rho^\varphi(x^{(i)},t)
\;\middle|\;
i=1,\ldots,50,\;
t\in\mathcal{T}_i
\right\}
\end{equation}
denote the set of robustness values obtained for specification $\varphi$
across all expert trajectories and valid evaluation times. We retain
\begin{equation}
\varphi \quad \text{iff} \quad Q_{0.50}(R_\varphi)\ge 0,
\end{equation}
i.e., the specification is removed if its median robustness is negative. Equivalently, a formula that is violated on more than half of the expert evaluation points is not used for reward construction. We refer to this step as expert-consistency filtering.

\begin{figure}[h]
\centering
\includegraphics[width=0.95\linewidth]{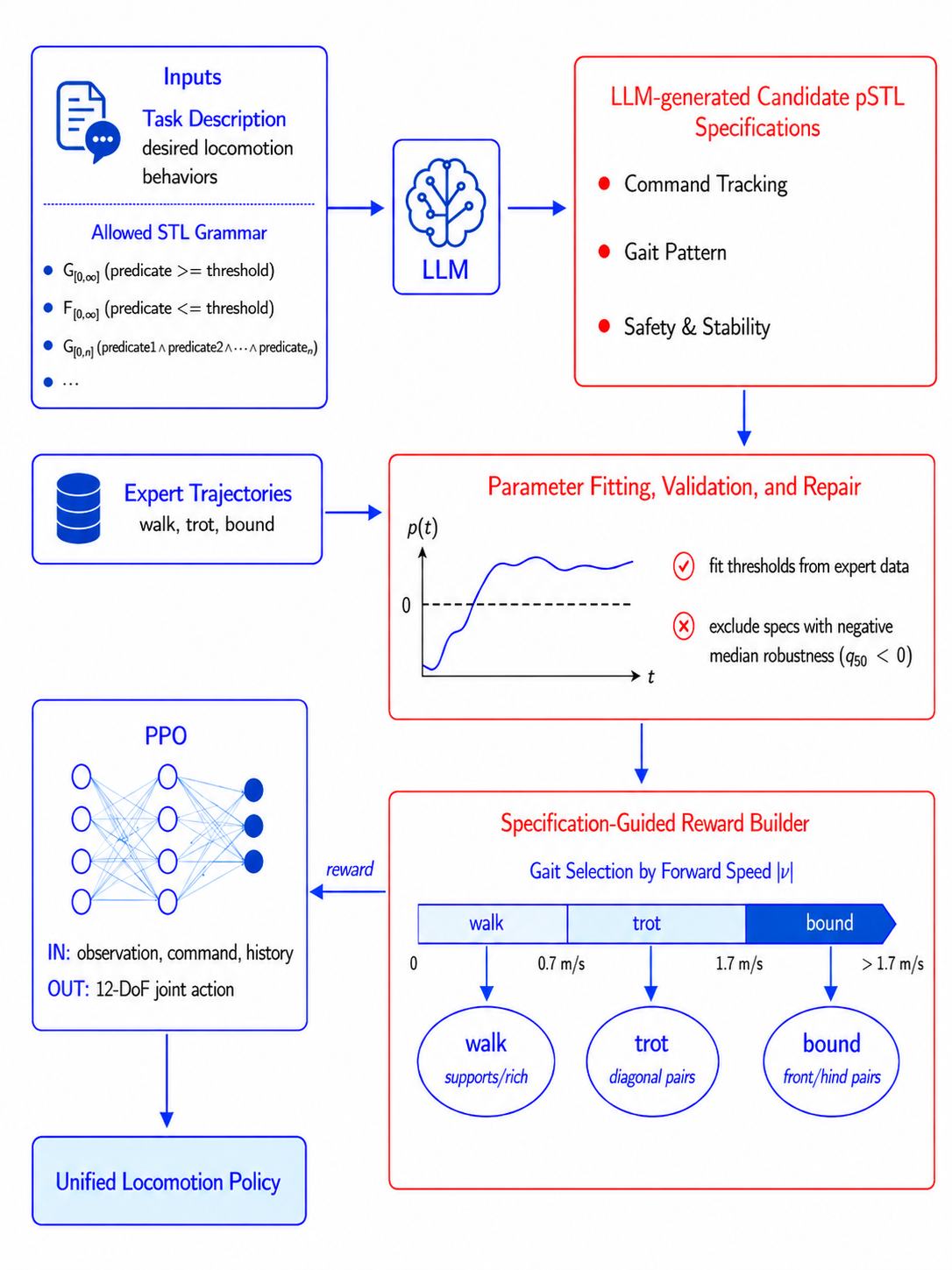}
\caption{Overall Pipeline.}
\label{fig:pipeline}
\end{figure}

\section{Experimental Design}

\subsection{Training and Baselines}

All learned policies use the same Barkour MJX environment, policy architecture, PPO implementation, command distribution, and domain-randomization scheme; only the reward source changes. We use friction randomization in $[0.6,1.4]$ and actuator gain perturbations, following the existing training setup. During training, a new velocity command is randomly sampled every 10 s. The initial command range is set to $v_y^{cmd} \in [-0.2,\,0.2]\,{m/s}$, $\omega^{cmd} \in [-0.2,\,0.2]\,{rad/s}$, and the forward-velocity range is gradually expanded to $v_x^{cmd} \in [0.0,\,1.9]\,{m/s}$ as part of the curriculum for learning the locomotion at higher speeds. Details of the training procedure and domain randomization are provided in the GitHub repository.

We compare the following methods: (1) \textbf{Heuristic-reward} uses the standard hand-engineered locomotion reward in the original Barkour paper \cite{caluwaerts2023barkour}, (2) \textbf{human-STL} is the manually structured STL reward from the previous work and serves as an ablation for whether the LLM can replace manual specification design, (3 \& 4) \textbf{GPT-STL} and \textbf{Qwen-STL} are each trained from the specifications generated by the corresponding LLM, (5) \textbf{Text2Reward} adapts direct LLM-generated reward code \cite{text2reward} (implemented using GPT-5.5 as the underlying language model), (6) \textbf{Expert-switching oracle} selects the corresponding specialized expert for the ground-truth commanded speed regime. It is not a learned unified controller and is reported as an oracle reference rather than a directly comparable training method.

\subsection{Evaluation}

We evaluate forward commands $v_x^{cmd}\in\{0.3,0.5,0.7,1.0, \\ 1.3,1.6,1.9,2.0,2.1\}$~m/s with zero lateral/yaw command, using 20 rollouts of 500 steps per command and excluding the first 50 steps for tracking metrics. We leverage four evaluation metrics: {\bf cost of transportation (CoT)}, {\bf survival rate}, {\bf success rate}, and {\bf gait match}. 

We measure energy efficiency using cost of transportation $\frac{E}{m\,g\,d}$ where $E$ is the total energy consumed, $m$ is the robot mass,
$g$ is gravitational acceleration, and $d$ is the planar distance traveled during the rollout. For episodes where the robot fails to survive, this metric is computed only up to the timestep of early termination. Lower CoT indicates better transport efficiency. 

A rollout is counted as \emph{survived} if the robot does not terminate early
due to falling or violating the environment termination conditions.

We define success based on velocity tracking accuracy after the warm-up phase.
Let $v_x^{cmd}$ be the commanded forward velocity, and let
$v_x^{loc}(t)$ denote the robot's forward velocity in its local body
frame. The average post-warm-up forward speed for rollout $k$ is
$
\bar{v}_x^{(k)}
=
\frac{1}{T-50}
\sum_{t=51}^{T}
v_x^{loc, k}(t).
$
A rollout is considered successful if
$
\left|\bar{v}_x^{(k)} - v_x^{cmd}\right|
\leq 0.15\,|v_x^{cmd}|,
$
i.e., the average forward speed stays within $\pm 15\%$ of the commanded
velocity. 

Gait match is a qualitative label based on the dominant contact pattern: Success denotes a dominant pattern matching the prescribed gait, Partial denotes intermittent matching, and Fail denotes a different dominant pattern.

For each LLM, robustness horizon $H$ is selected using validation performance from $\{1,5,10,20,30\}$.

\section{Results and Discussion}

Table~\ref{tab:gait_agnostic_results} shows that  GPT-GA and Text2Reward-GA (GA denotes gait-agnostic) achieve the strongest quantitative performance in the gait-agnostic setting. They maintain command success and survival across the full evaluated velocity range while yielding the lowest cost of transport (CoT) for most commands. In contrast, the heuristic-reward policy loses tracking accuracy at the two highest speeds, and Qwen-GA loses both survival and success from $1.6$ m/s onward. Despite the strong tracking performance of GPT-GA and Text2Reward-GA, visual inspection reveals a critical discrepancy between their quantitative results and their actual locomotion quality. The policies learn a bound-like contact pattern and exploits it across the entire tested speed range. At a low commanded velocity of $v_x^{cmd} = 0.3$ m/s, the robot continues to move its legs at a high cadence, producing pronounced vertical body oscillations. While this behavior yields a $100\%$ average velocity-tracking success rate, the resulting low-speed motion is visibly unnatural and undesirable. This highlights a fundamental limitation of evaluating locomotion solely through command success: high quantitative scores do not guarantee dynamically appropriate control. (Videos of the policies are provided on the project website).

This limitation underscores the value of the multi-gait formulation. When explicit gait structures are prescribed, the results in Table~\ref{tab:multigait_results} exhibit a markedly different pattern. Human-STL is the most consistent overall gait-aware controller. Qwen-MG matches its command/survival performance and desired bound behavior at high speed but fails to reproduce several prescribed lower-speed gaits. GPT-MG captures the low/mid-speed target gaits better but loses command tracking at 1.9 m/s and above. Text2Reward-MG is energy-efficient where it works but fails completely at high speeds. Ultimately, these results demonstrate that no single reward formulation universally dominates across both task settings and all evaluation criteria.

\textbf{GPT-5.5 vs Qwen 3.6:}
The relative performance of the two LLM-generated specification sets depends strongly on whether gait structure is prescribed. In the gait-agnostic setting, GPT-GA is more robust at high speeds: it survives all tested commands and tracks commands accurately, while Qwen-GA terminates from $1.6,\mathrm{m/s}$ onward. The comparison reverses in the multi-gait setting. Qwen-MG tracks commands throughout the full $0.3$-$2.1,\mathrm{m/s}$ range and generally matches or improves upon GPT-MG in CoT, whereas GPT-MG fails the command-success criterion at $1.9,\mathrm{m/s}$ and above. Thus, the quality of an LLM-generated specification cannot be characterized independently of the control setting in which it is used.

\textbf{Why the Text2Reward comparison matters:}
Both Text2Reward and our approach begin from language-model knowledge, but the intermediate representation differs. Text2Reward directly commits to numerical executable reward code from current-step signals. Our pipeline first exposes the proposed behavior as symbolic temporal constraints, then fits thresholds from expert data. This comparison lets us ask whether exposing the LLM output as an STL specification changes the learned behavior relative to directly generating reward code. 

The clearest difference between STL specification generation and direct reward-code generation appears in the high-speed regime. Text2Reward-MG terminates in every rollout at 1.9, 2.0, and 2.1 m/s, whereas Qwen-MG achieves 100\% survival and command success at all three commands and exhibits the desired bound contact pattern. This advantage is regime-specific: Qwen-MG does not reproduce the prescribed gait at several lower-speed commands. The comparison indicates that the STL intermediate representation can produce a more effective high-speed learning signal than the evaluated Text2Reward baseline, while also showing that successful command tracking does not guarantee recovery of all intended gait structures.

\begin{table}[t]
\centering
\scriptsize
\caption{
Gait-agnostic locomotion results across commanded forward velocities.
Lower CoT is better; higher survival and success are better. Each entry reports mean $\pm$ standard deviation over 20 rollouts.}
\label{tab:gait_agnostic_results}
\renewcommand{\arraystretch}{1.02}
\setlength{\tabcolsep}{6.5pt}
\begin{tabular}{|c c c c|}
\toprule
$v_x$ (m/s)
& CoT $\downarrow$
& Survival $\uparrow$
& Success $\uparrow$ \\
\midrule

\multicolumn{4}{l}{\textbf{Heuristic}} \\
0.3 & $1.20 \pm 0.00$ & $100\%$ & $100\%$ \\
0.5 & $1.00 \pm 0.00$ & $100\%$ & $100\%$ \\
0.7 & $1.00 \pm 0.00$ & $100\%$ & $100\%$ \\
1.0 & $1.20 \pm 0.00$ & $100\%$ & $100\%$ \\
1.3 & $1.30 \pm 0.00$ & $100\%$ & $100\%$ \\
1.6 & $1.40 \pm 0.00$ & $100\%$ & $100\%$ \\
1.9 & $1.40 \pm 0.00$ & $100\%$ & $100\%$ \\
2.0 & $1.40 \pm 0.00$ & $100\%$ & $5\%$ \\
2.1 & $1.40 \pm 0.00$ & $100\%$ & $0\%$ \\

\midrule
\multicolumn{4}{l}{\textbf{GPT-GA}} \\
0.3 & $1.16 \pm 0.01$ & $100\%$ & $100\%$ \\
0.5 & $1.02 \pm 0.01$ & $100\%$ & $100\%$ \\
0.7 & $0.99 \pm 0.01$ & $100\%$ & $100\%$ \\
1.0 & $0.95 \pm 0.01$ & $100\%$ & $100\%$ \\
1.3 & $0.95 \pm 0.01$ & $100\%$ & $100\%$ \\
1.6 & $0.95 \pm 0.01$ & $100\%$ & $100\%$ \\
1.9 & $0.94 \pm 0.01$ & $100\%$ & $100\%$ \\
2.0 & $0.95 \pm 0.01$ & $100\%$ & $100\%$ \\
2.1 & $0.97 \pm 0.01$ & $100\%$ & $100\%$ \\

\midrule
\multicolumn{4}{l}{\textbf{Qwen-GA}} \\
0.3 & $1.39 \pm 0.02$ & $100\%$ & $100\%$ \\
0.5 & $1.17 \pm 0.02$ & $100\%$ & $100\%$ \\
0.7 & $1.10 \pm 0.02$ & $100\%$ & $100\%$ \\
1.0 & $0.94 \pm 0.01$ & $100\%$ & $100\%$ \\
1.3 & $0.96 \pm 0.01$ & $100\%$ & $100\%$ \\
1.6 & $1.04 \pm 0.12$ & $0\%$ & $0\%$ \\
1.9 & $1.13 \pm 0.07$ & $0\%$ & $0\%$ \\
2.0 & $1.12 \pm 0.12$ & $0\%$ & $0\%$ \\
2.1 & $1.19 \pm 0.12$ & $0\%$ & $0\%$ \\

\midrule
\multicolumn{4}{l}{\textbf{Text2Reward-GA}} \\
0.3 & $0.91 \pm 0.01$ & $100\%$ & $100\%$ \\
0.5 & $0.80 \pm 0.00$ & $100\%$ & $100\%$ \\
0.7 & $0.75 \pm 0.00$ & $100\%$ & $100\%$ \\
1.0 & $0.74 \pm 0.00$ & $100\%$ & $100\%$ \\
1.3 & $0.79 \pm 0.00$ & $100\%$ & $100\%$ \\
1.6 & $0.87 \pm 0.00$ & $100\%$ & $100\%$ \\
1.9 & $1.00 \pm 0.01$ & $100\%$ & $100\%$ \\
2.0 & $1.07 \pm 0.01$ & $100\%$ & $100\%$ \\
2.1 & $1.14 \pm 0.01$ & $100\%$ & $100\%$ \\

\bottomrule
\end{tabular}
\end{table}

\textbf{Gait-aware vs gait-agnostic locomotion:}
The effect of explicitly prescribing gait structure is not uniform across methods. For Qwen-STL, the multi-gait formulation substantially extends the range of successful locomotion: Qwen-GA fails from $1.6,\mathrm{m/s}$ onward, while Qwen-MG maintains successful command tracking across the complete evaluation range. GPT-STL and Text2Reward show the opposite trend: their gait-agnostic policies succeed across the complete range, whereas the multi-gait version fails at the high-speed commands. These results suggest that adding gait-specific structure can improve high-speed locomotion, but its benefit depends on the particular reward or specification produced by the language model.

\textbf{Effect of robustness horizon:}
To understand the impact of the STL robustness evaluation horizon on policy learning, we conduct an ablation study over the window size $H \in \{1, 5, 10, 20, 30\}$ for both gait-agnostic and multi-gait formulations (Tables~\ref{tab:window_ablation_ga} and \ref{tab:window_ablation_mg}). Across most configurations, shorter temporal horizons ($H \in \{1, 5\}$) yield significantly more stable and successful locomotion policies. As the trailing window expands to $H=20$ and $H=30$, performance generally degrades. In the gait-agnostic setting, GPT-GA maintains a perfect $100\%$ success rate and optimal cost of transport (CoT $\le 1.00$) for $H \le 10$, but command tracking success collapses to $44\%$ at $H=20$. Similarly, in the multi-gait setting, Qwen-MG achieves perfect command tracking and high efficiency (CoT of $1.28$) at $H=1$, yet its performance degrades as the window size increases, dropping to a $27.8\%$ success rate at $H=30$. While GPT-MG exhibits a minor improvement in success rate at longer horizons in the multi-gait task, its overall performance remains well below Qwen-MG's peak. Furthermore, even when success and survival rates remain completely invariant to the horizon size, as seen with Qwen-GA, energy efficiency consistently worsens at longer horizons. These results show that a longer robustness horizon is not always beneficial. Because the robustness of $\mathbf{G}$ (always) specifications depends on the worst value in the window, larger $H$ can keep past violations in the reward longer and make credit assignment harder, whereas short horizons provide a more immediate learning signal.

\section{Conclusion}

We presented an LLM-assisted formal reward-design pipeline for quadruped locomotion. Instead of asking a language model to directly choose reward code and numerical coefficients, GPT-5.5 and Qwen 3.6 propose parametric STL structures; expert trajectories determine their parameters and filter out systematically violated suggestions; and smooth STL robustness provides the PPO reward. The resulting framework supports both gait-aware multi-gait control and a gait-agnostic setting in which locomotion style is allowed to emerge. These comparisons allow us to study what is gained and what remains difficult when temporal logic is inserted between language-model generation and policy learning.

A notable limitation is that our evaluations are currently confined to simulation (MJX). Despite employing domain randomization over friction and actuation parameters, the empirical sim-to-real transferability of these learned multi-gait behaviors has yet to be established on a physical quadruped.

\begin{table}[t]
\centering
\scriptsize
\caption{
Multi-gait locomotion results across commanded forward velocities.
Lower CoT is better; higher survival, command success, and gait success are better.
Gait success measures whether the desired gait associated with the commanded
speed regime is successfully exhibited.
}
\label{tab:multigait_results}
\renewcommand{\arraystretch}{1.02}
\setlength{\tabcolsep}{4.2pt}
\begin{tabular}{|c c c c c|}
\toprule
$v_x$
& CoT $\downarrow$
& Survival $\uparrow$
& Success $\uparrow$
& Gait Match $\uparrow$ \\
\midrule

\multicolumn{5}{l}{\textbf{Expert Oracle}} \\
0.3 & $0.92 \pm 0.02$ & $100\%$ & $100\%$  & Success \\
0.5 & $0.90 \pm 0.01$ & $100\%$ & $100\%$  & Success \\
0.7 & $0.96 \pm 0.01$ & $100\%$ & $100\%$  & Success \\
1.0 & $1.08 \pm 0.01$ & $100\%$ & $100\%$ & Success \\
1.3 & $1.27 \pm 0.01$ & $100\%$ & $100\%$ & Success \\
1.6 & $1.34 \pm 0.02$ & $100\%$ & $100\%$ & Success \\
1.9 & $1.33 \pm 0.01$ & $95\%$  & $0\%$ & NA \\
2.0 & $1.35 \pm 0.01$ & $95\%$  & $0\%$ & NA \\
2.1 & $1.36 \pm 0.01$ & $100\%$  & $0\%$ & NA \\

\midrule
\multicolumn{5}{l}{\textbf{Human-STL}} \\
0.3 & $2.10 \pm 0.10$ & $100\%$ & $100\%$ & Success \\
0.5 & $1.50 \pm 0.00$ & $100\%$ & $100\%$ & Success \\
0.7 & $1.20 \pm 0.00$ & $100\%$ & $100\%$ & Success \\
1.0 & $1.20 \pm 0.00$ & $100\%$ & $100\%$ & Success \\
1.3 & $1.10 \pm 0.00$ & $100\%$ & $100\%$ & Success \\
1.6 & $1.00 \pm 0.00$ & $100\%$ & $100\%$ & Success \\
1.9 & $1.10 \pm 0.00$ & $100\%$ & $100\%$ & Success \\
2.0 & $1.10 \pm 0.00$ & $100\%$ & $100\%$ & Success \\
2.1 & $1.10 \pm 0.00$ & $100\%$ & $100\%$ & Success \\

\midrule
\multicolumn{5}{l}{\textbf{GPT-MG}} \\
0.3 & $1.96 \pm 0.04$ & $100\%$ & $100\%$ & Success \\
0.5 & $1.62 \pm 0.03$ & $100\%$ & $100\%$ & Success \\
0.7 & $1.45 \pm 0.03$ & $100\%$ & $100\%$ & Success \\
1.0 & $1.27 \pm 0.02$ & $100\%$ & $100\%$ & Success \\
1.3 & $1.17 \pm 0.01$ & $100\%$ & $100\%$ & Success \\
1.6 & $1.26 \pm 0.01$ & $100\%$ & $100\%$ & Partially \\
1.9 & $1.27 \pm 0.01$ & $100\%$ & $0\%$ & NA \\
2.0 & $1.28 \pm 0.01$ & $100\%$ & $0\%$ & NA \\
2.1 & $1.28 \pm 0.02$ & $100\%$ & $0\%$ & NA \\

\midrule
\multicolumn{5}{l}{\textbf{Qwen-MG}} \\
0.3 & $1.97 \pm 0.05$ & $100\%$ & $100\%$ & Fail \\
0.5 & $1.42 \pm 0.03$ & $100\%$ & $100\%$ & Fail \\
0.7 & $1.22 \pm 0.02$ & $100\%$ & $100\%$ & Fail \\
1.0 & $1.18 \pm 0.01$ & $100\%$ & $100\%$ & Fail \\
1.3 & $1.10 \pm 0.01$ & $100\%$ & $100\%$ & Fail \\
1.6 & $1.09 \pm 0.01$ & $100\%$ & $100\%$ & Success \\
1.9 & $1.15 \pm 0.02$ & $100\%$ & $100\%$ & Success \\
2.0 & $1.19 \pm 0.01$ & $100\%$ & $100\%$ & Success \\
2.1 & $1.22 \pm 0.01$ & $100\%$ & $100\%$ & Success \\

\midrule
\multicolumn{5}{l}{\textbf{Text2Reward-MG}} \\
0.3 & $1.10 \pm 0.02$ & $100\%$ & $100\%$ & Success \\
0.5 & $0.82 \pm 0.01$ & $100\%$ & $100\%$ & Success \\
0.7 & $0.83 \pm 0.01$ & $100\%$ & $100\%$ & Success \\
1.0 & $1.04 \pm 0.01$ & $100\%$ & $100\%$ & Fail \\
1.3 & $1.01 \pm 0.01$ & $100\%$ & $100\%$ & Fail \\
1.6 & $0.98 \pm 0.01$ & $100\%$ & $100\%$ & Fail \\
1.9 & $1.05 \pm 0.15$ & $0\%$ & $0\%$ & NA \\
2.0 & $1.06 \pm 0.18$ & $0\%$ & $0\%$ & NA \\
2.1 & $1.05 \pm 0.17$ & $0\%$ & $0\%$ & NA \\

\bottomrule
\end{tabular}
\end{table}

\begin{table}[t]
\centering
\scriptsize
\caption{
Ablation of STL robustness window size $H$ for gait-agnostic policies. Metrics are averaged across all evaluated commanded forward velocities.
}
\label{tab:window_ablation_ga}
\setlength{\tabcolsep}{7.0pt}
\renewcommand{\arraystretch}{1.05}

\begin{tabular}{@{}|l |c | c c c|@{}}
\toprule
Method & $H$ & CoT $\downarrow$ & Survival $\uparrow$ & Success $\uparrow$ \\
\midrule

\multirow{5}{*}{GPT-GA}
& 1  & 0.99 & 100\% & 100\% \\
& 5  & 0.99 & 100\% & 100\% \\
& 10 & 1.00 & 100\% & 100\% \\
& 20 & 1.15 & 100\% & 44\% \\
& 30 & 1.14 & 100\% & 67\% \\

\midrule

\multirow{5}{*}{Qwen-GA}
& 1  & $1.11$ & $55.6\%$ & $55.6\%$ \\
& 5  & $1.11$ & $55.6\%$ & $55.6\%$ \\
& 10 & $1.26$ & $55.6\%$ & $55.6\%$ \\
& 20 & $1.29$ & $55.6\%$ & $55.6\%$ \\
& 30 & $1.55$ & $55.6\%$ & $55.6\%$ \\

\bottomrule
\end{tabular}
\end{table}

\begin{table}[t]
\centering
\scriptsize
\caption{
Ablation of STL robustness window size $H$ for multi-gait policies. Metrics are averaged across all evaluated commanded forward velocities.
}
\label{tab:window_ablation_mg}
\setlength{\tabcolsep}{7.0pt}
\renewcommand{\arraystretch}{1.05}

\begin{tabular}{@{}|l | c | c c c |@{}}
\toprule
Method & $H$ & CoT $\downarrow$ & Survival $\uparrow$
& Success $\uparrow$ \\
\midrule

\multirow{5}{*}{GPT-MG}
& 1  & 1.40 & 100\% & 55.6\%  \\
& 5  & 1.60 & 100\% & 55.6\%  \\
& 10 & 1.37 & 100\% & 55.6\%  \\
& 20 & 1.40 & 100\% & 66.7\%  \\
& 30 & 1.36 & 99.4\% & 66.7\%  \\

\midrule

\multirow{5}{*}{Qwen-MG}
& 1  & $1.28$ & $100\%$ & $100\%$ \\
& 5  & $1.32$ & $100\%$ & $77.8\%$  \\
& 10 & $2.08$ & $88.3\%$ & $60.6\%$ \\
& 20 & $14.25$ & $77.8\%$ & $54.4\%$ \\
& 30 & $3.59$ & $77.8\%$ & $27.8\%$ \\

\bottomrule
\end{tabular}
\end{table}

\bibliographystyle{IEEEtran}
\bibliography{ref}

\end{document}